\documentclass[10pt,twocolumn,letterpaper]{article}

\usepackage{wacv} 
\usepackage{graphicx}
\usepackage{amsmath}
\usepackage{amssymb}
\usepackage{booktabs}
\usepackage{amssymb,amsmath}
\usepackage{textgreek}
\usepackage{balance}
\usepackage{xcolor}

\usepackage[pagebackref,breaklinks,colorlinks]{hyperref}

\usepackage[capitalize]{cleveref}
\crefname{section}{Sec.}{Secs.}
\Crefname{section}{Section}{Sections}
\Crefname{table}{Table}{Tables}
\crefname{table}{Tab.}{Tabs.}

\def\wacvPaperID{7} 
\def\confName{WACV}
\def\confYear{2025}

\begin{document}

\title{TextureCrop: Enhancing Synthetic Image Detection through Texture-based Cropping}

\author{Despina Konstantinidou\\
CERTH - ITI\\
Thessaloniki, Greece\\
{\tt\small dekonstantinidou@iti.gr}
\and
Christos Koutlis\\
CERTH - ITI\\
Thessaloniki, Greece\\
{\tt\small ckoutlis@iti.gr}
\and
Symeon Papadopoulos\\
CERTH - ITI\\
Thessaloniki, Greece\\
{\tt\small papadop@iti.gr}
}
\maketitle

\begin{abstract}
    Generative AI technologies produce increasingly realistic imagery, which, despite its potential for creative applications, can also be misused to produce misleading and harmful content. This renders Synthetic Image Detection (SID) methods essential for identifying AI-generated content online. State-of-the-art SID methods typically resize or center-crop input images due to architectural or computational constraints, which hampers the  detection of artifacts that appear in high-resolution images. To address this limitation, we propose TextureCrop, an image pre-processing component that can be plugged in any pre-trained SID model to improve its performance. By focusing on high-frequency image parts where generative artifacts are prevalent, TextureCrop  enhances SID performance with manageable memory requirements. Experimental results demonstrate a consistent improvement in AUC across various detectors by 6.1\% compared to center cropping and by 15\% compared to resizing, across high-resolution images from the Forensynths, Synthbuster and TWIGMA datasets. Code available at \url{https://github.com/mever-team/texture-crop}.
\end{abstract}

\section{Introduction}
\label{sec:intro}

\begin{figure*}[tb]
    \begin{subfigure}[b]{0.135\textwidth}
        \includegraphics[width=\textwidth]{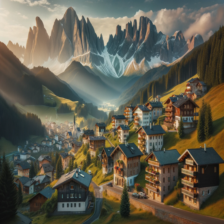}
        \caption{}
        \label{fig:resized}
    \end{subfigure}
    \hfill
    \begin{subfigure}[b]{0.135\textwidth}
        \includegraphics[width=\textwidth]{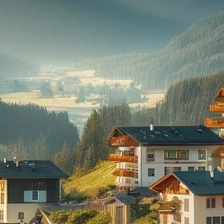}
        \caption{}
        \label{fig:centercropped}
    \end{subfigure}
    \hfill
    \begin{subfigure}[b]{0.34\textwidth}
        \includegraphics[width=\textwidth]{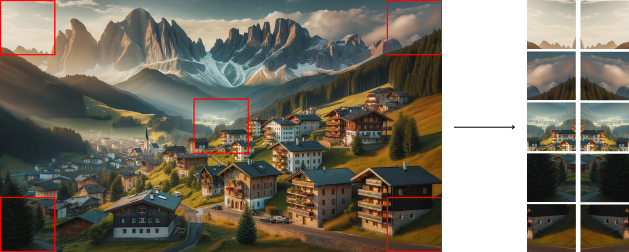}
        \caption{}
        \label{fig:ten_crop}
    \end{subfigure}
    \hfill
    \begin{subfigure}[b]{0.34\textwidth}
        \includegraphics[width=\textwidth]{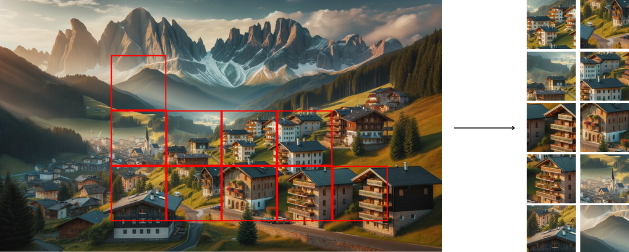}
        \caption{}
        \label{fig:texture_crop}
    \end{subfigure}
    \caption{Example of the crops created from a 1792x1024 image by using (a) resizing, (b) center cropping, (c) ten cropping and (d) texture cropping. Images (a) and (b) have the same dimensions ($224\times224$) as the 10 crops shown in (c) and (d).}
    \label{fig:all_images}
\end{figure*}

In recent years, the research field of synthetic media generation has seen tremendous growth, as attested by the emergence of  numerous state-of-the-art methods~\cite{jabbar2020survey,wang2021generative,dhariwal2021diffusion,song2021score,yang2023diffusion,croitoru2022diffusion}. Generative Adversarial Networks (GANs)~\cite{goodfellow2014gan} and Diffusion models~\cite{ho2020diffusion} have revolutionized the field of synthetic image generation by enabling the creation and editing of images in very high quality and fidelity. While these technologies offer immense creative potential, they also present serious risks for misuse, such as spreading disinformation, facilitating impersonation and fraudulent activities, compromising biometric security systems and carrying out foreign interference campaigns. The quality of generated images has reached a level where even the most careful observers can be deceived~\cite{luFake2M, papa2023ontheuse}, posing a significant challenge for the majority of Internet users. Moreover, the diversity of fake image sources complicates the detection process, necessitating the development of robust and universal detection methods.

Researchers have recently dedicated significant effort to developing solutions to  the problem. Synthetic Image Detection (SID) is typically performed based on pixel- or frequency-based features. Some methods apply extensive data augmentation~\cite{wang2020cnngenerated,gragnaniello2021gan,chen2022ost} in order to improve generalization, while others exploit the traces left by the generation process~\cite{frank2020leveraging,jeong2022frepgan,quian2020thinking}. Recently, numerous  methods~\cite{amoroso2023parents,ojha2023universal,sha2023defake,cozzolino2023raising,koutlis2024leveraging} have relied on features extracted by foundational models like CLIP~\cite{radford2021clip}, which have shown remarkable effectiveness in distinguishing between real and generated images.

With the advancement of generative AI models, there has been remarkable improvement in the ability to generate high-resolution and high-quality images, which are widely shared across the Web. Most state-of-the-art detection models perform exceptionally well on low-resolution images, but consistently underperform on high-resolution images. This challenge arises because the majority of these detectors are trained using substantially smaller images (in most cases $256\times256$ pixels). Typically, these detectors adapt images to fixed input layers using resizing or center cropping. However, resizing entails image resampling and interpolation, which may erase the subtle high-frequency traces left by the generation process~\cite{corvi2022detection} (see \Cref{fig:resized}) and center cropping an image to a fraction of its original size can lead to significant loss of information (see \Cref{fig:centercropped}). When detectors are trained and evaluated on small images, resizing or cropping has little effect on their performance since only few high-frequency traces are lost, but when dealing with images of higher resolution, this effect is much more pronounced. 

To address this issue, we introduce TextureCrop, an easily adaptable pre-processing SID component specifically designed for high-resolution images. The proposed method targets high-frequency image patches that encapsulate fine details and textures crucial for distinguishing between synthetic and real images. It leverages a sliding window technique to systematically analyze image patches and adopts a sampling mechanism that retains patches exhibiting significant texture information while discarding the rest. We conduct an extensive comparison among standard pre-processing techniques during inference to showcase the effectiveness of the proposed method. Experimental results on high-resolution images demonstrate substantial improvements in detection metrics. 
Specifically, TextureCrop demonstrates an increase in balanced accuracy ranging from 3.4\% to 12.1\%, an improvement in area under the ROC curve ranging from 6.1\% to 14.9\%, and an enhancement in average precision ranging from 3.3\% to 18.2\% across a variety of state-of-the-art SID detectors on more than 10,000 high-resolution real and generated images from the Forensynths~\cite{wang2020cnngenerated}, Synthbuster~\cite{bammey2023synthbuster}, TWIGMA~\cite{chen2023twigmadatasetaigeneratedimages}, RAISE~\cite{ductien2015raise} and OpenImages~\cite{openimages} datasets.

\subsection{Overview}

\begin{figure*}[tb]
    \centering
    \includegraphics[width=\textwidth]{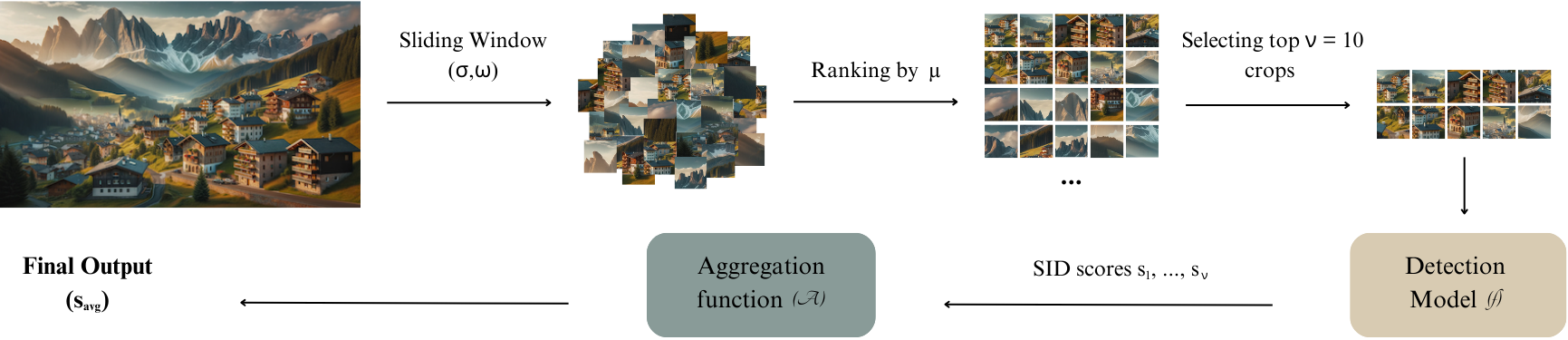}
    \caption{Overview of the TextureCrop Pipeline.}
    \label{fig:framework}
\end{figure*}

Our main contributions can be summarized as follows:
\begin{enumerate}
    \item We introduce TextureCrop, an image pre-processing method aimed at enhancing the detection of high-resolution synthetic images by focusing on high-frequency texture components. 
    \item We conduct a comprehensive analysis of various pre-processing techniques applied to high-resolution images, exploring their impact on detection performance. Our findings demonstrate that TextureCrop consistently outperforms existing pre-processing approaches across various detectors.
    \item We present a detailed ablation study to explore the parameters of the proposed TextureCrop method. 
\end{enumerate}


\section{Related Work}
\subsection{Synthetic image generation}

Over the past few years, image generation models have consistently improved in creating high-quality realistic images. Most early works focused on Generative Adversarial Networks (GAN), and their numerous extensions~\cite{mirza2014cgan,radford2016dcgan,zhu2017cyclegan,karras2018progan,choi2018stargan,brock2019biggan,karras2019stylegan,park2019gaugan}.The seminal StyleGAN architecture~\cite{karras2019stylegan} and its subsequent versions, namely  StyleGAN2~\cite{karras2020analyzing} and StyleGAN3~\cite{karras2021alias}, introduced a style-based approach with adaptive instance normalization, enabling fine-grained control over image attributes and improved disentanglement of the latent space for generating highly realistic and diverse images.

More recently, Diffusion Models (DM)~\cite{nichol2021glide,rombach2022latent,ramesh2022dalle2,betker2023dalle3,rombach2022stable,podell2023sdxl,adobe2023firefly,midjourney2023} have emerged as an alternative approach to image generation. Notable among these are text-to-image generation models like DALL-E 2~\cite{ramesh2022dalle2} and DALL-E 3~\cite{betker2023dalle3}. DALL-E 2 employs a diffusion process where noise is progressively added to the image during training to corrupt it, creating a series of increasingly noisy versions. During generation, the model reverses this process, starting from pure noise and iteratively denoises it step-by-step, guided by the latent representation of the input text prompt. DALL-E 3 builds upon its predecessor with improved training methods and dataset utilization. Stable diffusion models, such as Stable Diffusion 2~\cite{rombach2022stable} and Stable Diffusion XL~\cite{podell2023sdxl}, utilize latent representations from training data to progressively add noise to images. Firefly~\cite{adobe2023firefly}, developed by Adobe, generates images by iteratively refining a noise-filled input through a diffusion process, guided by text prompts and user-defined parameters to create high-quality, customizable outputs.

Beyond GANs and DMs, there are also generative models focused on pixel-level manipulation and enhancement. For example, Second Order Attention Network (SAN)~\cite{dai2019san} enhances single image super-resolution using second-order attention mechanisms and recursive feature networks. Similarly, Seeing In The Dark (SITD)~\cite{chen2018sitd} employs fully convolutional networks to simulate long-exposure photography from low-light conditions based on short-exposure raw camera input.

\subsection{Synthetic image detection}
Following the rapid developments in image generation, developing robust deep learning-based SID methods has also attracted ample research interest. Training set diversity has been found to help generalize to unseen architectures, as shown in~\cite{wang2020cnngenerated},  where a pre-trained ResNet50~\cite{kaiming2016resnet} is trained on 20 different object classes of real images from LSUN~\cite{yu2015lsun} and ProGAN~\cite{karras2018progan} images. Avoiding resizing has also been found to be important~\cite{corvi2022detection}, as it may erase the artifacts left by the generative model. In order to preserve these subtle high-frequency traces, some methods propose working on local patches~\cite{chai2020patchforensics}, as well as analyzing both local and global features~\cite{ju2022fusing}. Other methods propose avoiding downsampling steps in the first layers of the network~\cite{gragnaniello2021gan,corvi2022detection}. 

Recently, there have been efforts to leverage the capabilities of pre-trained vision-language models (VLMs) for the SID task~\cite{amoroso2023parents,ojha2023universal,sha2023defake,cozzolino2023raising}. For instance, Ohja et al.~\cite{ojha2023universal} use the same training protocol as in~\cite{wang2020cnngenerated}, but employ the Contrastive Language Image Pre-Training (CLIP) model~\cite{radford2021clip}, as a feature extractor instead of training a ResNet50. Another method~\cite{koutlis2024leveraging} proposes the extraction of features not only from the last layer of CLIP’s image encoder, but also from intermediate layers. 

Moreover, recent works have focused on analyzing the texture characteristics of images. It has been observed that synthetic faces exhibit significantly different textures compared to real ones~\cite{liu2020global}. Additionally,~\cite{zhong2024patchcraft} highlights that texture patches contain more distinct traces left by generative models than global semantic information, while~\cite{chen2024single} demonstrates that generative models tend to focus on generating the patches with rich textures to make the images more realistic and neglect the hidden noise caused by camera that is present in simple patches. 

Recognizing the potential of rich texture information that is contained in local image regions, 
we propose a pre-processing method that selectively identifies and prioritizes such regions to be used in their native resolution by SID models. 


\section{Proposed Approach}

\subsection{Problem Formulation}
\label{subsec:problem_formulation}
The goal of SID methods is to determine whether an image, $\mathrm{I} \in \mathbb{R}^{K\times W\times H}$, is generated or real, where $C$, $H$, and $W$ denote the number of channels (in most cases 3), height and width of the image, respectively. In the typical formulation, a single image $\mathrm{I}$ (or a batch of images), is fed into the model $f$, resulting in a SID score $s \in [0, 1]$, indicating the likelihood of the image being synthetic:

\begin{equation}
s = f(\mathrm{I})
\end{equation}

The majority of SID methods often assume that a resized or cropped version of an image is processed, which may result in significant loss of information. Few detectors~\cite{gragnaniello2021gan, corvi2022detection} process the whole image in its native resolution, but in that case processing time and memory requirements increase proportionally to the image's size. TextureCrop addresses this issue by first dividing image $\mathrm{I}$ into a set of smaller crops $\{ \mathrm{C}_i \}_{i=1}^{N_I} \in \mathbb{R}^{N_I\times K \times W^c \times H^c}$ using a sliding window approach, where $\mathrm{C}_i$ represents the i-th crop with dimensions $C \times W^c \times H^c$ and $N_I$ being the number of crops generated for image $I$, which depends on the dimensions of the original image. Subsequently, the crops are ranked according to their texture information (cf. \Cref{subsec:selecting}), and the top $\nu$ crops are selected for further analysis. Each crop $\mathrm{C}_i$ is processed through the same detection model $f$, returning a SID score $\mathrm{s}_i$. The final SID score $s$ is computed by aggregating the individual scores from the selected crops using an aggregation function $\mathfrak{A}$:
\begin{align}
& \{ \mathrm{C}_i \}_{i=1}^{\nu} = \text{TextureCrop}(\mathrm{I}) \\
& s_i = f(\mathrm{C}_i), \quad \text{for } i = 1, \ldots, \nu \\
& s = \mathfrak{A}(\{ s_i \}_{i=1}^{\nu}) 
\end{align}

\subsection{Selecting texture-rich image crops}\label{subsec:selecting}

Employing a sliding window approach with user-defined stride $\sigma$ and window size $\omega$, where $\omega$ = 224 to align with standard input dimensions for pretrained deep learning models, TextureCrop systematically traverses the input image $\mathrm{I}$ and extracts the corresponding crops $\{ \mathrm{C}_i \}_{i=1}^{N_I}$ similarly to convolving the image. For images with dimensions that are not exact multiples of the window size, the method ensures complete coverage by including additional overlapping crops. Specifically, when the sliding window reaches the edge of the image along either the width or height, and the remaining area is less than the window size, the algorithm selects crops starting from the last valid position while overlapping with previously selected regions. More complex methods, such as Region Proposal Networks (RPNs)~\cite{ren2016fasterrcnnrealtimeobject}, were also tested; however, they showed no improvement in detection accuracy despite their significantly increased processing time. 

Each crop $\mathrm{C}_i$ is then converted to grayscale, and its texture information is measured based on a measure $\mu$. The method then sorts the crops based on this measure and selects the top $\nu$ crops. The resulting cropped regions 
\[
\{ \mathrm{C}_i \}_{i=1}^{\nu} = \text{TextureCrop}(I, \sigma, \omega, \nu, \mu, p),
\]
are then aggregated using an appropriate method $\mathfrak{A}$, and a final prediction is generated.

\subsection{Parameters}

This subsection outlines the key parameters $\sigma, \omega, \nu, \mu$ used in TextureCrop. Each of these parameters controls different aspects of the cropping behavior. In this section, we mention the proposed values for these parameters, while their detailed evaluation and experimentation with alternative values is presented in \cref{subsec:ablations}. 

\subsubsection{Stride ($\sigma$)}

The stride parameter, $\sigma$, controls the step size of the sliding window. For the experiments reported in this work, we selected a stride of 224, to strike a balance between exploring multiple regions of the image and ensuring computational efficiency. 

\subsubsection{Window Size ($\omega$)}

The window size parameter, $\omega$, represents the size of the crops generated by the sliding window. Most methods in our analysis use a fixed input size of $224\times 224$ pixels, which is why we set $\omega$ to this value, as it is a common choice for pretrained models. Exceptions include GANID~\cite{gragnaniello2021gan} (ProGAN and StyleGAN2) and DMID~\cite{corvi2022detection}, which support a variety of input sizes.

\subsubsection{Number of Crops ($\nu$)}

The number of crops parameter, $\nu$, specifies how many crops are selected from the image. While increasing $\nu$ allows for more robust analysis by providing better coverage of the input image, it also leads to higher processing time and computational requirements. After experimentation, we set $\nu = 10$ as the default value.

\subsubsection{Texture Selection Measure ($\mu$)}

TextureCrop selects texture-rich regions of the image using a specific measure $\mu$, namely the Global Histogram Entropy (GHE)~\cite{sezgin2004new}.GHE provides a measure of the overall randomness in intensity distribution within each crop. After the image is converted to grayscale, the histogram of pixel intensities is computed and normalized to the range $[0, 1]$ and divided into $n$ = 256 bins, where the frequency of pixel intensities in each bin is recorded as $h_j$. This is then normalized to obtain a probability distribution:
\[p_j = \frac{h_j}{\sum_{k=1}^{n} h_k}, \quad \text{for } j = 1, \dots, n,\]
ensuring that $\sum_{j=1}^{n} p_j = 1$. The entropy $\varepsilon_i$ is then calculated using Shannon's formula:
\[
\varepsilon_i = -\sum_{j=1}^{n} p_j \log_2(p_j + \epsilon),
\]
where $\epsilon = 10^{-7}$ is a small constant added to avoid undefined values for $\log_2(0)$. 

While GHE is the best performing measure used in TextureCrop, other alternatives can also be employed, including measures such as Local Entropy (LE), Standard Deviation (SD), Inverse Autocorrelation (\(\text{AC}^{-1}\)) and Texture Diversity (TD), which can capture different aspects of texture information. We explored these alternative measures in the ablations section (see~\Cref{subsubsec:measure}) to evaluate their impact on performance. 



\subsection{Aggregation method}

Our method proposes the processing of a pre-defined number of crops per image, creating the need for an aggregation method to consolidate the scores from individual crops into a single prediction. The best performing aggregation method was found to be the logits' $\{ s_i \}_{i=1}^{\nu}$ average:
\[
s_{\text{avg}} = \mathfrak{A}(\{ s_i \}_{i=1}^{\nu})=\frac{1}{\nu} \sum_{i=1}^\nu s_i,
\]
where $s_i$ denotes the $i$-th crop's logit.

\begin{table*}[tb]
  \caption{Forensynths, Synthbuster and TWIGMA datasets after removing low resolution images.}
  \label{tab:dataset}
  \resizebox{\textwidth}{!}{%
  \begin{tabular}{lccccc}
  \toprule
  Subset & Dataset & Total images & Real images & Generated images & Resolutions produced by the generative model\\
  \midrule
  SAN & Forensynths & 20 & 10 & 10 & depends on the resolution of the input image \\
  SITD & Forensynths & 360 & 180 & 180 & depends on the resolution of the input image \\
  Whichfaceisreal & Forensynths & 2000 & 1000 & 1000 & up to $1024\times1024$ \\
  DALL-E 2 & Synthbuster & 1000 & * & 1000 &  $256 \times256$, $512\times512$, $1024\times1024$ \\
  DALL-E 3 & Synthbuster & 1000 & * & 1000 & $1024\times1024$, $1024\times1792$, $1792\times1024$ \\
  Firefly & Synthbuster & 1000 & * & 1000 & $2048\times2048$, $2304\times1792$, $1792\times2304$, $2688\times1536$ \\
  Stable Diffusion 2 & Synthbuster & 41 & * & 41 & $512\times512$, $768\times768$\\
  Stable Diffusion XL & Synthbuster & 41 & * & 41 & $1024\times1024$, $1152\times896$, $896\times1152$, $1216\times832$, and others \\
  TWIGMA & TWIGMA & 5000 & ** & 5000 & various \\
  \bottomrule\\
  \end{tabular}%
  }
  \tiny
  * For real data, we used 1000 high-resolution images from the RAISE dataset.\\
  ** For real data, we used 1093 high-resolution images from the OpenImages dataset.
\end{table*}

\section{Experimental Setup}

\subsection{Image Pre-processing Approaches for SID}
\label{subsec:imgpre}

Contemporary detectors commonly utilize resizing or cropping as pre-processing steps. Resizing adjusts an image to a specified target size through scaling, a process that can introduce distortions and diminish fine details, particularly when altering images to significantly divergent dimensions. Cropping is typically implemented through CenterCrop, i.e. selecting a square or rectangular area from the center of an image, or TenCrop, i.e. extracting 10 distinct crops from an image (its four corners, central region, and their horizontal flips, ~\cref{fig:ten_crop}). 

\subsection{Datasets}

We evaluate our method using high-resolution images sourced from three datasets: Forensynths~\cite{wang2020cnngenerated}, Synthbuster~\cite{bammey2023synthbuster}, and TWIGMA~\cite{chen2023twigmadatasetaigeneratedimages}. After filtering out images with resolutions lower than $1024\times1024$, we collected a total of 2,380 real and generated images from the Forensynths dataset, 3,082 generated images from the Synthbuster dataset, and all 1,000 uncompressed and unprocessed high-resolution real images from the RAISE dataset~\cite{ductien2015raise}. Additionally, we included a subset of 5,000 generated images from the TWIGMA dataset in our evaluation, along with 1,093 high-resolution real images from the OpenImages~\cite{openimages} dataset, selected also based on the criterion of having a resolution of $1024\times1024$ pixels or greater. \Cref{tab:dataset} presents an overview of these datasets.

More specifically, the Forensynths dataset includes high-resolution images generated by the SAN~\cite{dai2019san} and SITD~\cite{chen2018sitd} models, as well as faces generated by the StyleGAN~\cite{karras2019stylegan} model, available on the WhichFaceIsReal website~\cite{whichfaceisreal}. The Forensynths dataset also includes real images for comparison. For the Synthbuster dataset, high-resolution images were generated by DALL-E 2~\cite{ramesh2022dalle2}, DALL-E 3~\cite{betker2023dalle3}, Firefly~\cite{adobe2023firefly}, Stable Diffusion 2~\cite{rombach2022stable}, and Stable Diffusion XL~\cite{podell2023sdxl}. The subset of the TWIGMA dataset that we used comprises 5,000 AI-generated images sourced from Twitter between January 2021 and March 2023, representing a diverse set of images generated by models including DALL-E 2~\cite{ramesh2022dalle2}, Midjourney~\cite{midjourney2023}, and Stable Diffusion~\cite{rombach2022stable}.

\subsection{SID Detectors}

We include the following state-of-the-art pre-trained detectors in our study: GramNet~\cite{liu2020global} that leverages global image texture representations, CNNDetect~\cite{wang2020cnngenerated} a ResNet50 with blurring and compression augmentation (with two different probabilities of the image being augmented: 0.1 and 0.5), GANID~\cite{gragnaniello2021gan} a ResNet50 with intense augmentation that avoids down-sampling in the first layer (trained on ProGAN or StyleGAN2 data), DMID~\cite{corvi2022detection}, which proposes the fusion of the logits of two models same as described before but one trained on GAN and the other on diffusion data, UnivFD~\cite{ojha2023universal} that uses linear probing on features extracted from CLIP’s ViT-L/14 image encoder~\cite{dosovitskiy2020transformers}, RINE~\cite{koutlis2024leveraging} that uses the image representations extracted by intermediate layers of CLIP’s image encoder (trained on 4 object classes of ProGAN data and on Latent Diffusion Model (LDM) data and PatchCraft~\cite{zhong2024patchcraft}, which exploits the inter-pixel correlation contrast between rich and poor texture regions within an image. To conduct our experiments, we used the SIDBench framework~\cite{schinas2024sidbenchpythonframeworkreliably}. The checkpoints for all detectors were obtained from the original repositories of the respective papers.

\subsection{Evaluation metrics}

To evaluate the performance of the proposed pre-processing technique, we compute balanced accuracy (BA), average precision (AP), and area under the ROC curve (AUC) metrics. Each of these addresses different aspects of SID performance. 
The results presented are averages across all datasets, aiming at a robust evaluation of the method's performance. 


\section{Results}
\subsection{Performance Gains}

\begin{table*}[tb]
  \caption{Performance (BA/AP/AUC) of detection methods on high resolution images using different pre-processing methods.}
  \label{tab:metrics}
  \resizebox{\textwidth}{!}{%
  \begin{tabular}{lccccc}
    \toprule
    BA / AP / AUC  & Resizing & CenterCrop & TenCrop & TextureCrop & Performance Difference \\
    \midrule
    GramNet & 
    48.74 / 42.16 / 48.71 & 49.76 / 50.63 / 48.39 & 46.19 / 52.45 / 46.14 & \textbf{52.44} / \textbf{53.82} / \textbf{52.63} & 
    \textcolor{green}{+2.68} / \textcolor{green}{+1.36} / \textcolor{green}{+3.92} \\    
    CNNDetect (0.1) & 
    54.09 / 48.09 / 55.45 & 59.32 / 56.48 / 61.36 & 59.89 / 56.86 / 57.72 & \textbf{62.11} / \textbf{62.89} / \textbf{70.11} & 
    \textcolor{green}{+2.22} / \textcolor{green}{+6.03} / \textcolor{green}{+8.75}\\
    CNNDetect (0.5) & 
    51.80 / 50.86 / 57.09 & 54.82 / 52.80 / 56.43 &  \textbf{57.90} / 54.18 / 52.70 & 57.17 / \textbf{58.64} / \textbf{63.75} & 
    \textcolor{red}{-0.73} / \textcolor{green}{+4.46} / \textcolor{green}{+6.67} \\
    GANID (ProGAN) & 55.00 / 49.67 / 57.02 & 
    60.82 / 58.76 / 66.64  & 60.69 / 59.61 / 65.56 & \textbf{64.68} / \textbf{65.61} / \textbf{74.06} & 
    \textcolor{green}{+3.85} / \textcolor{green}{+6.00} / \textcolor{green}{+7.41} \\
    GANID (StyleGAN2) & 57.87 / 53.91 / 59.70 & 64.79 / 64.38 / 73.43 & 66.34 / 70.06 / 75.33 & \textbf{68.33} / \textbf{71.94} / \textbf{78.66} &
    \textcolor{green}{+1.99} / \textcolor{green}{+1.88} / \textcolor{green}{+3.32} \\
    
    DMID & 
    57.38 / 54.21 / 63.75 & 75.07 / 80.98 / 81.95 & 76.18 / 84.45 / 81.72 & \textbf{78.74} / \textbf{86.78} / \textbf{85.63} & 
    \textcolor{green}{+2.56} / \textcolor{green}{+2.33} / \textcolor{green}{+3.91} \\    
    UnivFD & 
    59.36 / 59.94 / 66.48 & 63.89 / 63.86 / 70.31 & 64.93 / \textbf{71.87} / 72.82 & \textbf{69.68} / 71.78 / \textbf{75.96} & 
    \textcolor{green}{+4.75} / \textcolor{red}{-0.09} / \textcolor{green}{+3.14} \\    
    RINE (4) & 
    63.15 / 62.75 / 69.21 & 69.55 / 69.97 / 73.44 & 69.32 / 74.60 / 74.25 & \textbf{71.24} / \textbf{76.66} / \textbf{78.27} & 
    \textcolor{green}{+1.69} / \textcolor{green}{+2.05} / \textcolor{green}{+4.02} \\    
    RINE (LDM) & 
    59.38 / 59.45 / 68.05 & 73.97 / 82.13 / 81.90 & \textbf{81.79} / 88.49 / \textbf{86.17} & 80.46 / \textbf{88.70} / 86.02  & 
    \textcolor{red}{-1.32} / \textcolor{green}{+0.21} / \textcolor{red}{-0.15} \\
    
    PatchCraft & 
    47.06  /  42.83  /  46.58 & 63.04  /  56.49  /  65.90 & 60.88  /  59.80  /  63.32 & \textbf{73.04}  /  \textbf{68.93}  /  \textbf{75.53} & 
    \textcolor{green}{+10.00}  /  \textcolor{green}{+9.14} / \textcolor{green}{+9.64} \\    \bottomrule
  \end{tabular}%
  }
\end{table*}

In this section, we evaluate the effectiveness of TextureCrop in enhancing SID performance across various models. We compare texture cropping against the standard pre-processing techniques of \cref{subsec:imgpre}, namely Resizing, CenterCrop and TenCrop.

On average across all detectors and datasets, TextureCrop exhibits an increase of 12.1\% increase in BA, 18.2\% in AP and 14.9\% in AUC compared to Resizing. Compared to CenterCrop, it offers a 4.3\% increase in BA, 6.9\% in AP and 6.1\% in AUC across the 10 methods tested (see~\Cref{tab:metrics}). Compared to TenCrop, it exhibits a 3.4\% increase in BA, 3.3\% in AP and 6.5\% in AUC. Among the methods tested, PatchCraft demonstrates the largest gains with a 10.00\% increase in BA, 9.14\% in AP, and 9.64\% in AUC, highlighting its strong performance with TextureCrop. On the other hand, RINE LDM shows the smallest improvements, with a slight decrease of 1.32\% in BA, a small increase of 0.21\% in AP, and a negligible decrease of 0.15\% in AUC.

CenterCrop and Resizing consistently display worse performance on average across the datasets compared to TextureCrop, across all methods. In contrast, TenCrop is the only method that occasionally outperforms TextureCrop, but even in these cases, it is only by a slight margin. The results demonstrate a high degree of consistency across all datasets except for the DALL-E 3 images from Synthbuster, where Resizing seems to lead to slightly better results. 

It is also worth noting that, compared to the methods that process the entire image, TextureCrop significantly reduces computational overhead. For example, for a $2048\times2048$ image, TextureCrop reduces processing time from 1.05s to 0.63s and lowers memory usage from 17 GB to 2.08 GB. Additionally, it addresses the limitations of methods that process the entire image, which are often constrained by memory and computational restrictions, as these methods can only handle images up to a certain size.

\subsection{Ablation Study}
\label{subsec:ablations}
In this section, we present the ablation studies conducted in order to assess the impact of the different parameters on the effectiveness of TextureCrop. 

\subsubsection{Stride}

\begin{figure}[tb]
    \begin{subfigure}[b]{0.15\textwidth}
        \includegraphics[width=\textwidth]{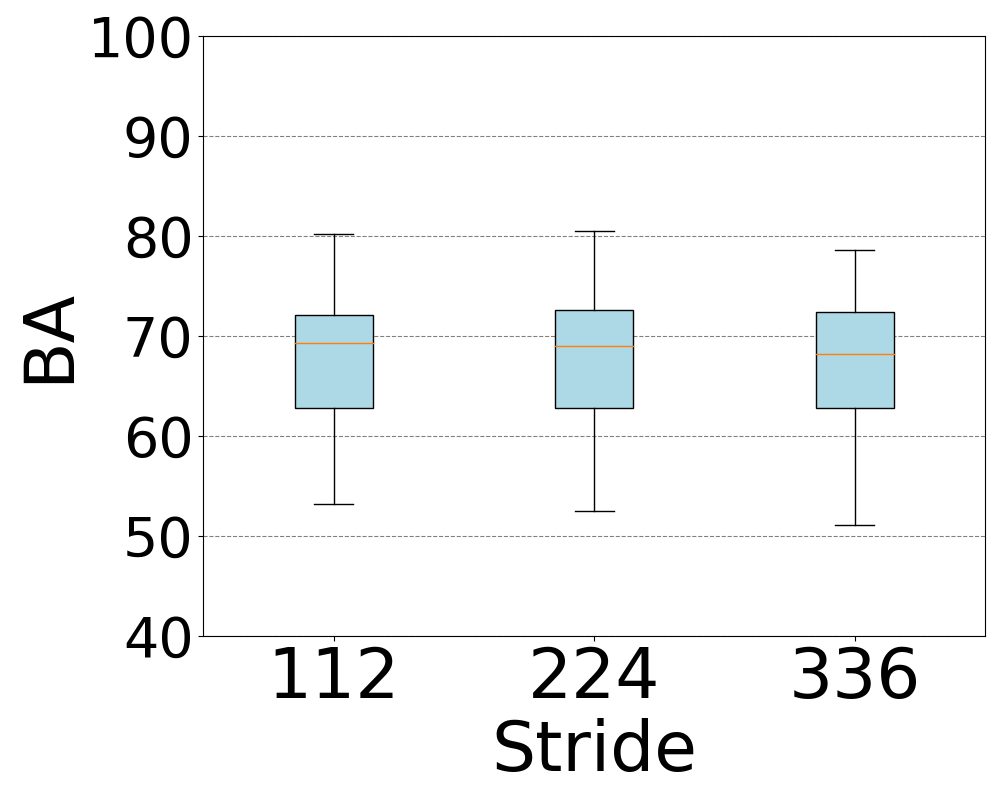}
        \caption{}
        \label{fig:stride-ba}
    \end{subfigure}
    \hfill
    \begin{subfigure}[b]{0.15\textwidth}
        \includegraphics[width=\textwidth]{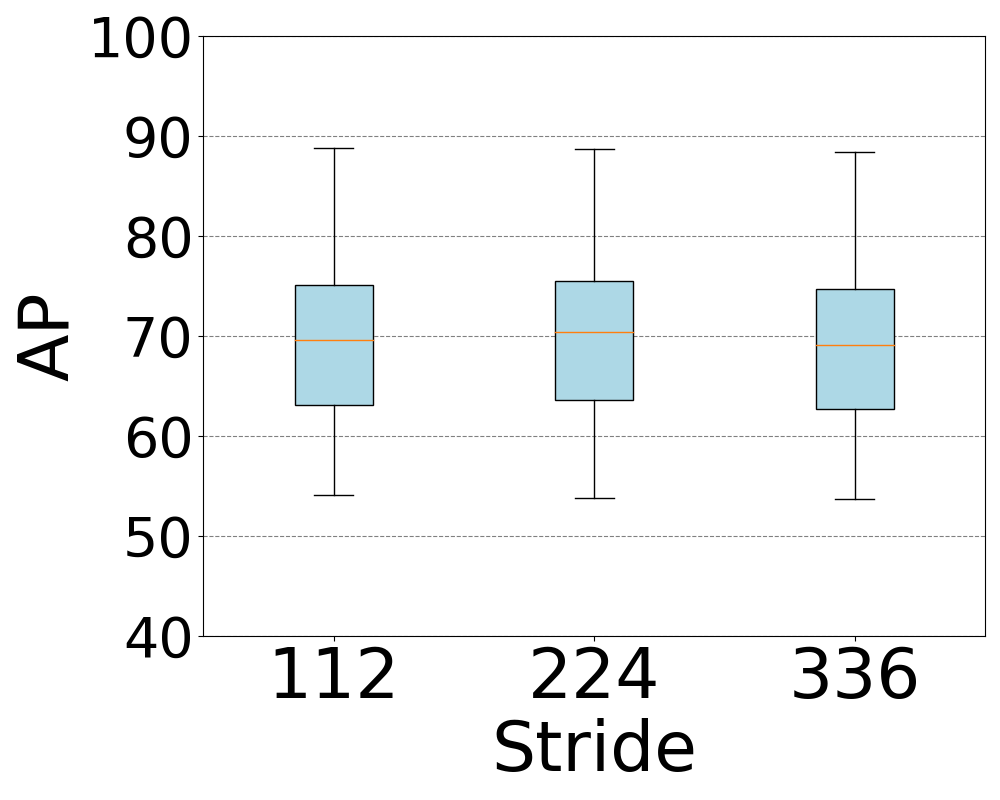}
        \caption{}
        \label{fig:stride-ap}
    \end{subfigure}
    \hfill
    \begin{subfigure}[b]{0.15\textwidth}
        \includegraphics[width=\textwidth]{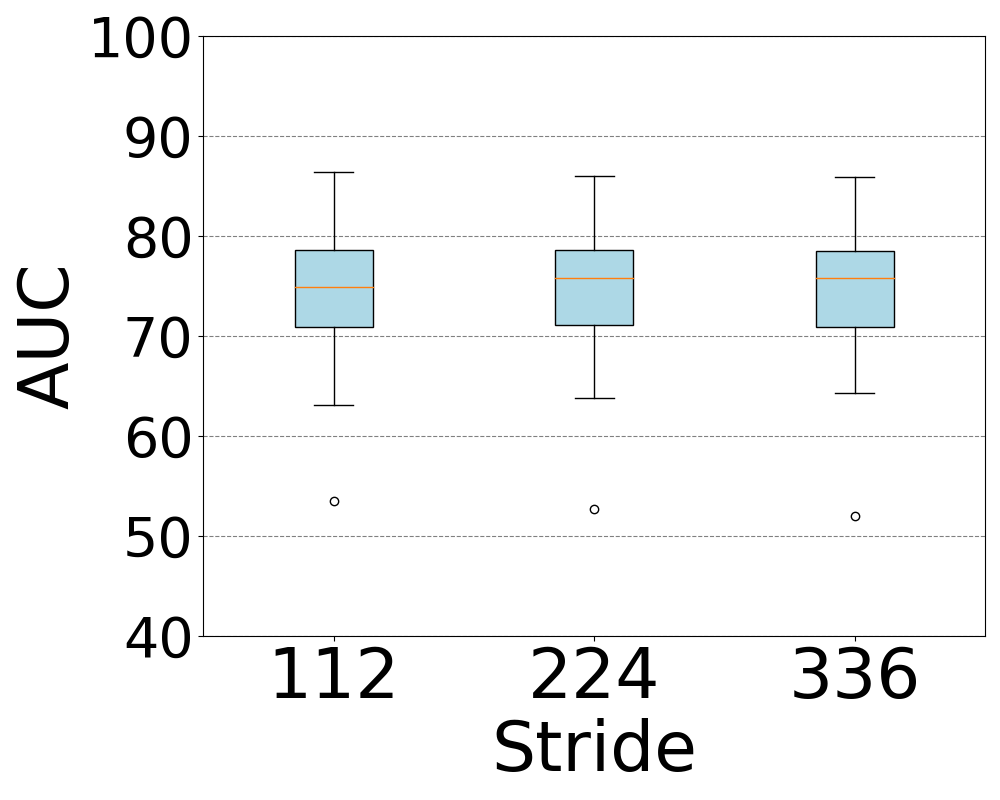}
        \caption{}
        \label{fig:stride-auc}
    \end{subfigure}
    \hfill
    \begin{subfigure}[b]{0.15\textwidth}
        \includegraphics[width=\textwidth]{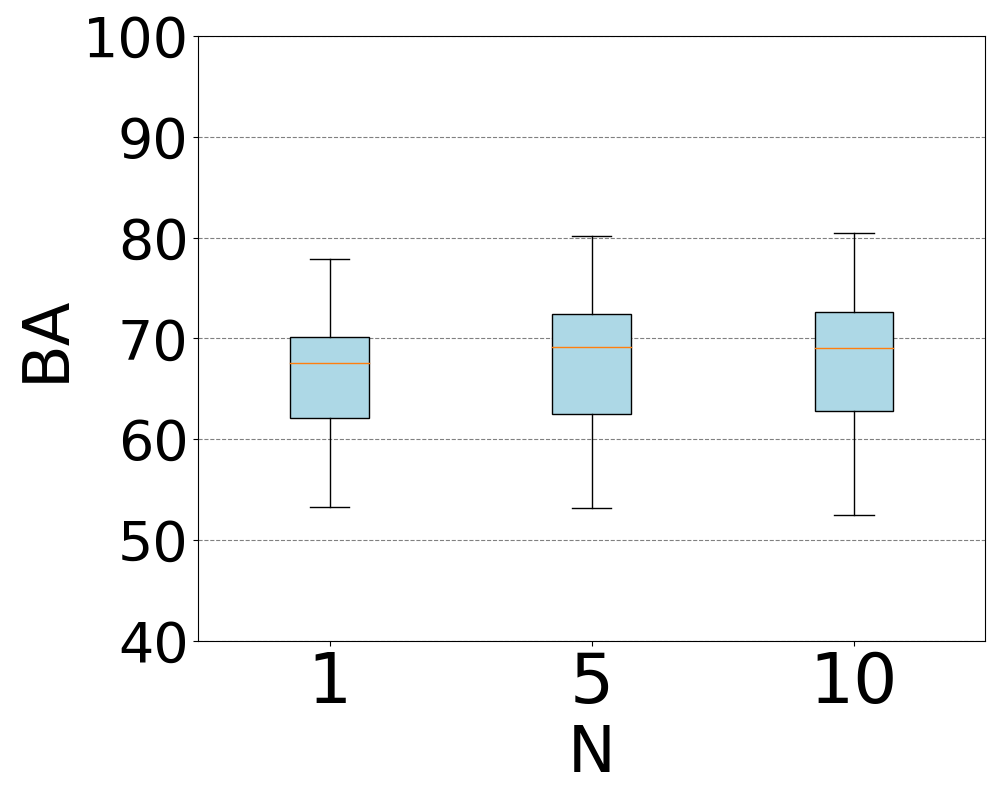}
        \caption{}
        \label{fig:n-ba}
    \end{subfigure}
    \hfill
    \begin{subfigure}[b]{0.15\textwidth}
        \includegraphics[width=\textwidth]{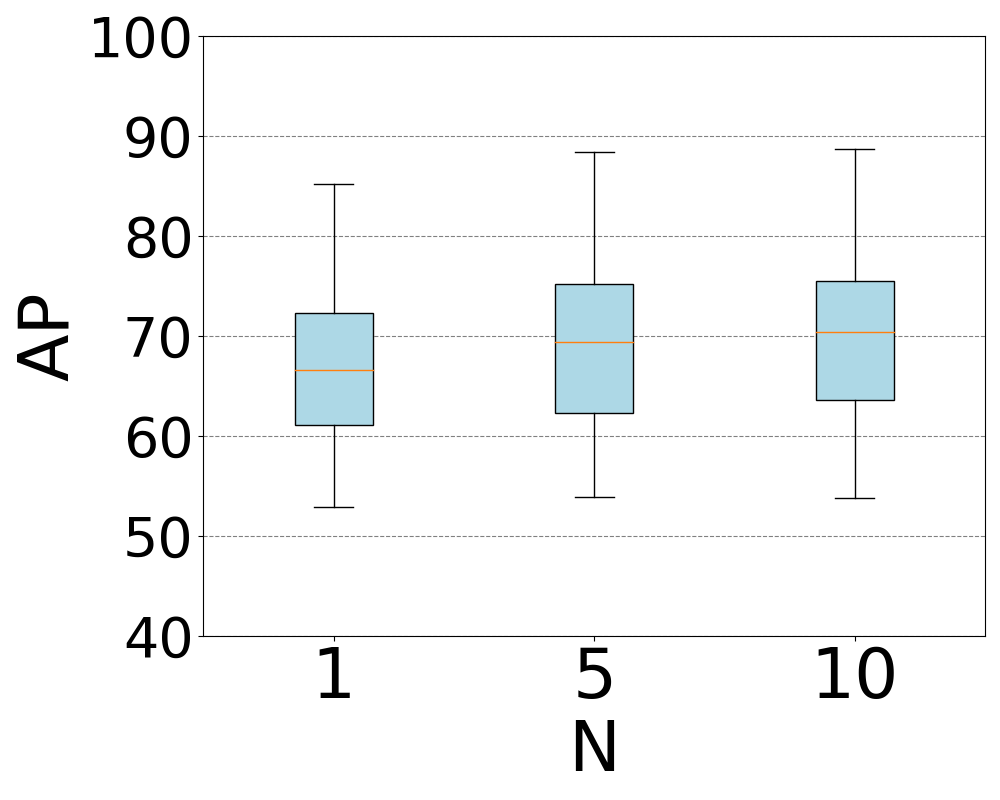}
        \caption{}
        \label{fig:n-ap}
    \end{subfigure}
    \hfill
    \begin{subfigure}[b]{0.15\textwidth}
        \includegraphics[width=\textwidth]{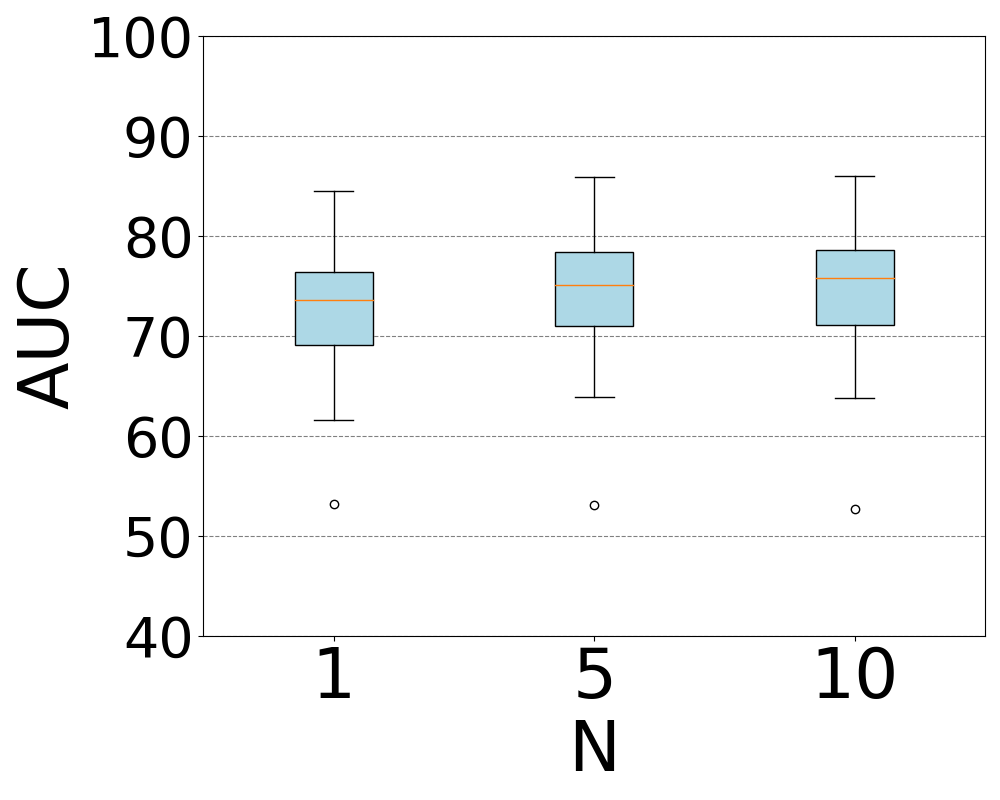}
        \caption{}
        \label{fig:n-auc}
    \end{subfigure}
    \caption{Box plot of (a) BA, (b) AP and (c) AUC distribution across detection methods for different values of the stride parameter $\sigma$, and (d) BA, (e) AP and (f) AUC distribution for different values of the number of crops parameter $\nu$.}
    \label{fig:ablations}
\end{figure}

We experimented with strides $\sigma$ of 112, 224, and 336 pixels to evaluate the trade-off between computational efficiency and detection performance. Our experiments revealed that the method’s performance is relatively insensitive to this parameter, yielding comparable results across all three stride values, as shown in~\Cref{fig:stride-ba,fig:stride-ap,fig:stride-auc}. Although no significant differences were observed, a stride $\sigma$ of 224 pixels is selected, as it ensures sufficient coverage of the image.

\subsubsection{Window Size}

Most methods in our analysis have a set input size of $224\times224$ pixels, except for GANID (ProGAN and StyleGAN2) and DMID. To accommodate these variations, we tested our method with the window size parameter set to $224\times224$, $512\times512$, and $1024\times1024$ pixels to determine the optimal size for capturing high-frequency artifacts while maintaining computational efficiency. Our findings indicate that a window size of $224\times224$ pixels provides the best results, performing on average 1-3\% better than $512\times512$ and nearly 1\% better than $1024\times1024$ across performance metrics.

\subsubsection{Number of Crops}

We examine the relationship between the number of crops $\nu$, and the effectiveness of TextureCrop. 
Results are presented in~\Cref{fig:n-ba,fig:n-ap,fig:n-auc} and reveal a noticeable decrease in performance when only one crop is selected compared to selecting 5 or 10 crops. While the results for 5 and 10 crops are similar, there is a slight performance increase with 10 crops. Despite this increase, processing 10 crops for a high resolution image is not computationally intensive. As such, the associated computational overhead remains modest, and is justified by the performance improvement observed with 10 crops.

In addition to this approach, we also tested another method that selects crops based on a thresholding criterion applied to the entropy values. This alternative strategy allows us to include only those crops with entropy values above a specified threshold. While both methods aim to capture texture-rich areas, the fixed number of crops approach demonstrated better performance and is more straightforward to implement, particularly for batch processing, as it consistently produces a fixed number of images for further analysis. Even when using the best-performing threshold value, the thresholding-based method resulted in 1-2\% lower performance across BA, AP and AUC compared to the fixed number of crops approach.

\subsubsection{Texture Selection Measure}
\label{subsubsec:measure}

\begin{figure}[tb]
    \begin{subfigure}[b]{0.23\textwidth}
        \includegraphics[width=\textwidth]{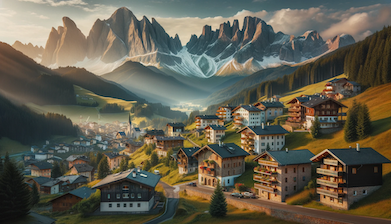}
        \caption{}
        \label{fig:image}
    \end{subfigure}
    \hfill
    \begin{subfigure}[b]{0.23\textwidth}
        \includegraphics[width=\textwidth]{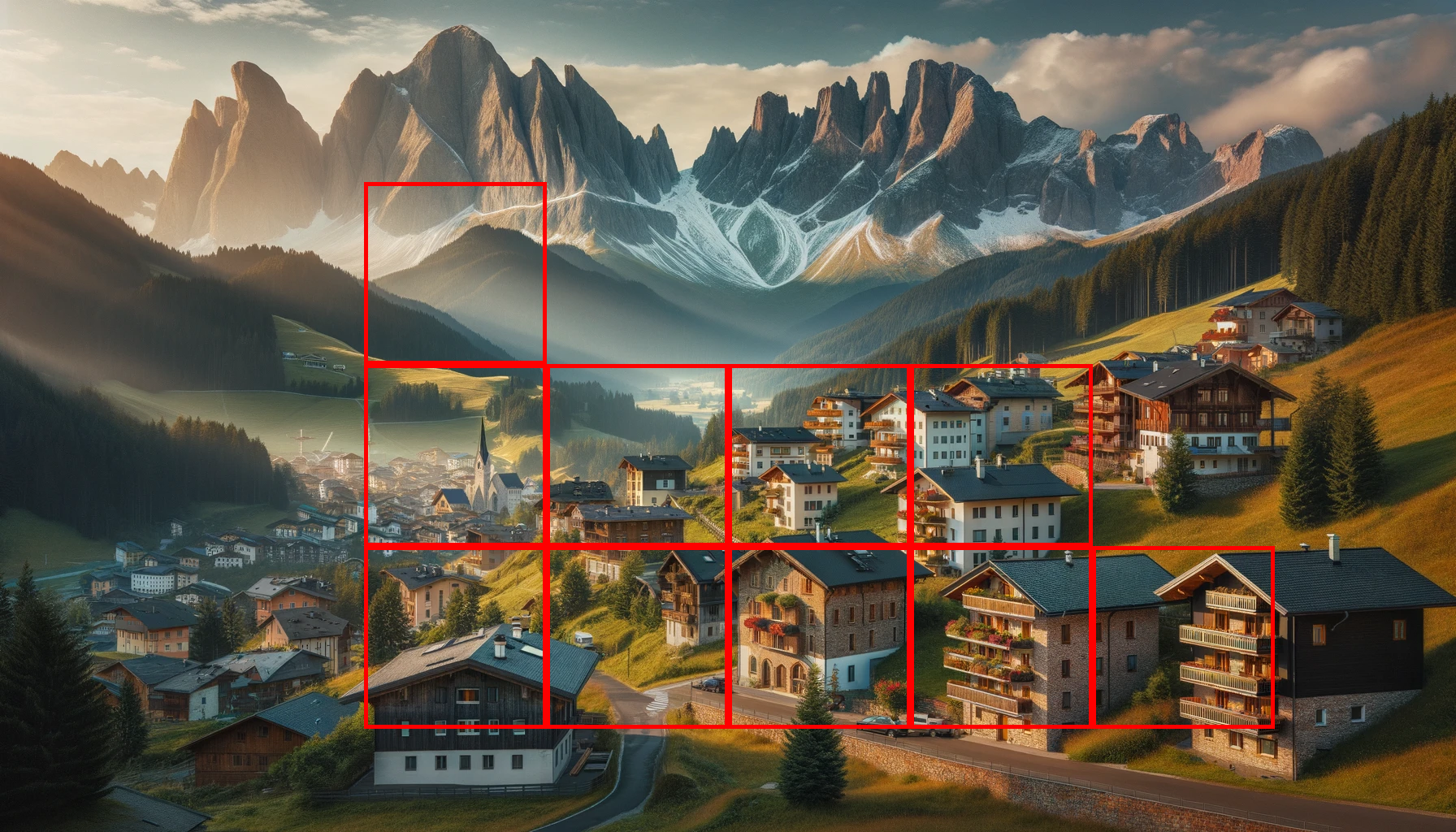}
        \caption{}
        \label{fig:texturecropped}
    \end{subfigure}
    \hfill
    \begin{subfigure}[b]{0.23\textwidth}
        \includegraphics[width=\textwidth]{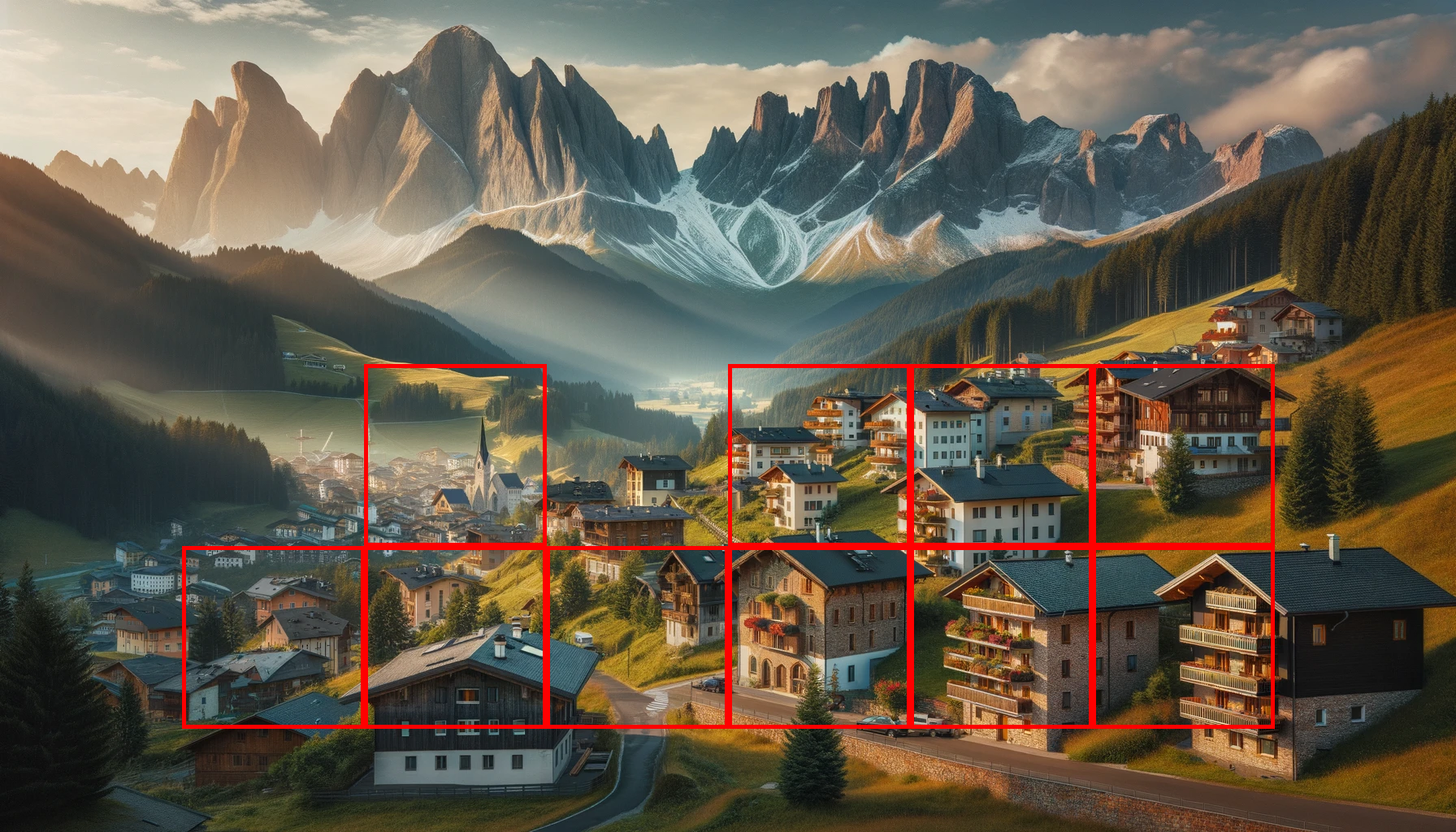}
        \caption{}
        \label{fig:entropy}
    \end{subfigure}
    \hfill
    \begin{subfigure}[b]{0.23\textwidth}

        \includegraphics[width=\textwidth]{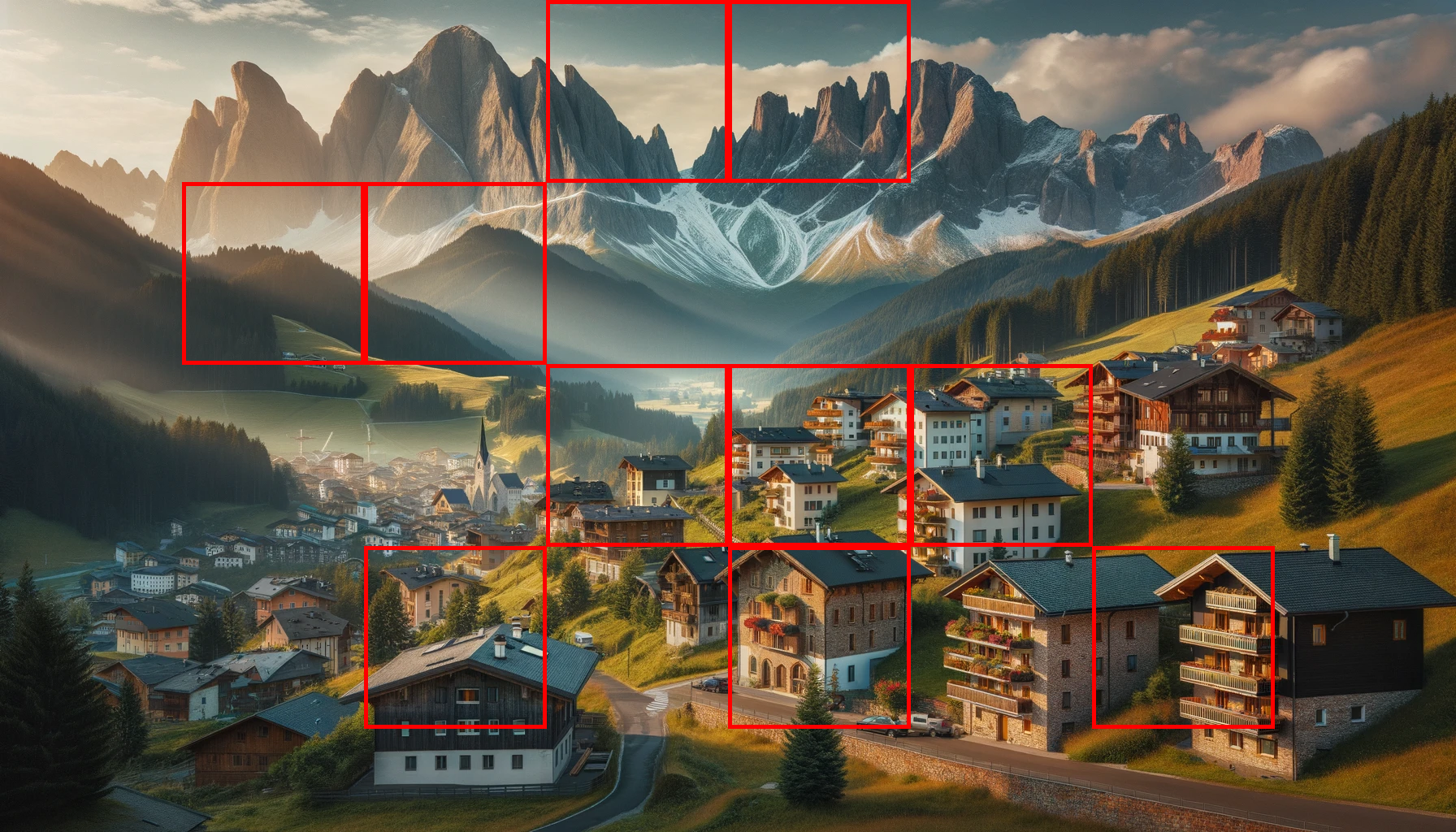}
        \caption{}
        \label{fig:sd}
    \end{subfigure}
    \hfill
    \begin{subfigure}[b]{0.23\textwidth}
        \includegraphics[width=\textwidth]{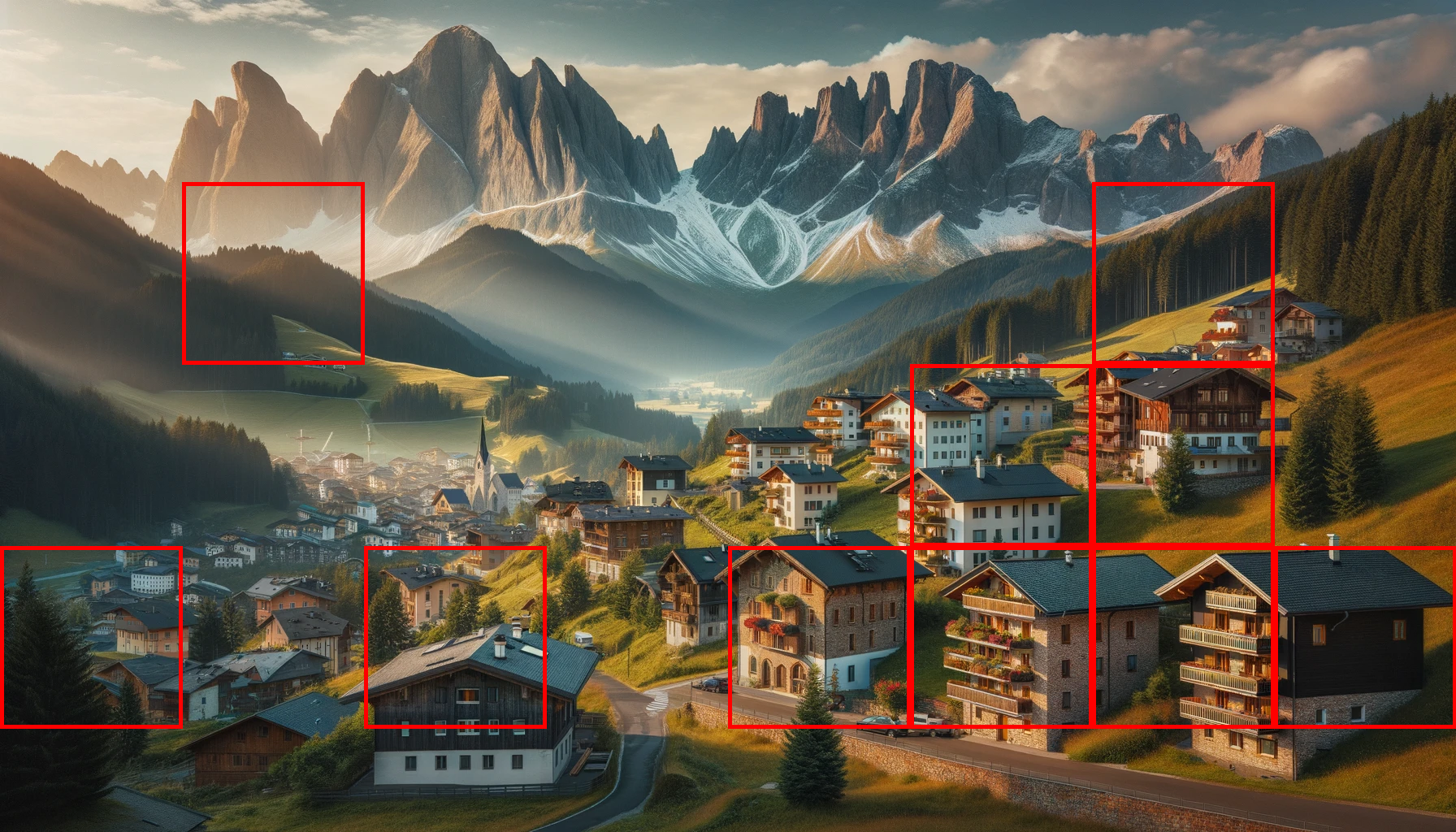}
        \caption{}
        \label{fig:autocorrelation}
    \end{subfigure}
    \hfill
    \begin{subfigure}[b]{0.23\textwidth}
        \includegraphics[width=\textwidth]{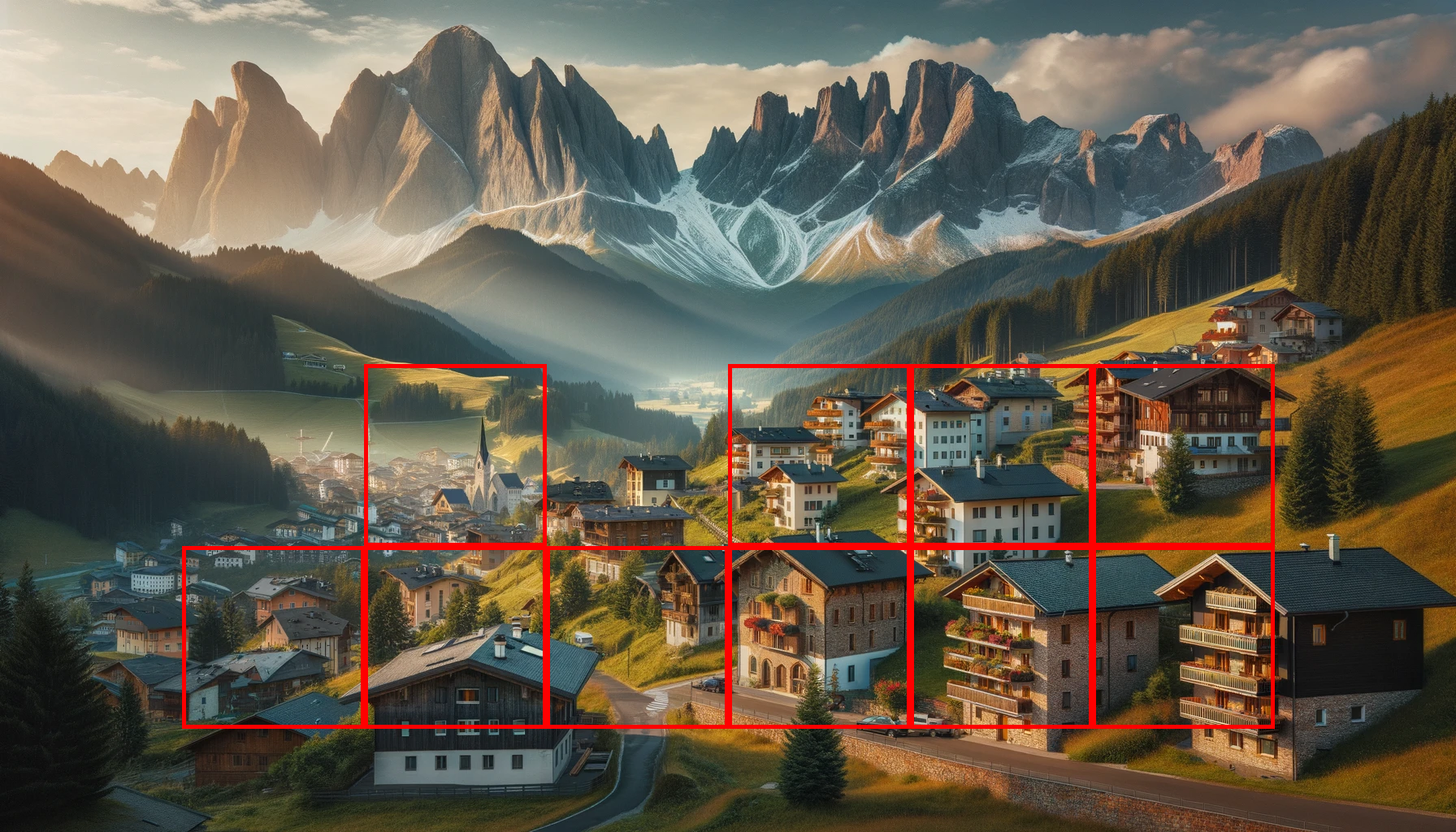}
        \caption{}
        \label{fig:texture_diversity}
    \end{subfigure}
    \caption{Example of the crops created from a 1792x1024 image (a) using texture cropping with global histogram entropy (b), local entropy (c), standard deviation (d), inverse autocorrelation (e) and texture diversity (f) as a metric.}
    \label{fig:metrics}
\end{figure}

The TextureCrop method allows for flexibility, supporting alternative measures $\mu$ besides GHE, including:

\begin{itemize}
    \item\textbf{Local Entropy} (LE) is derived by applying a local entropy filter across small neighborhoods within the crop. 
    Specifically, the entropy of each pixel in the crop is calculated using a local neighborhood filter, typically a disk-shaped window with a defined radius (in our implementation a disk filter of size 10 pixels). The resulting entropy values are then averaged across the entire crop.
    \item\textbf{Standard Deviation} (SD) of pixel intensities provides a measure of contrast and variability within a crop by calculating the variability of pixel intensities around their mean.  \item\textbf{Autocorrelation} (AC) is calculated by shifting the crop and assessing the correlation between the original and shifted version. Lower autocorrelation values may suggest the presence of distinct textural features. 
    \item\textbf{Texture Diversity} (TD), initially proposed in~\cite{zhong2024patchcraft}, quantifies the variations in pixel intensities across the crop by calculating the absolute differences between adjacent pixels in horizontal, vertical, and diagonal directions. 
\end{itemize}
These measures are applied to each crop individually to assess texture information. As we can see in~\Cref{fig:metrics}, in most cases, different measures result in different crops being selected, highlighting the importance of choosing an appropriate one for crop selection. These are designed to rely on simple mathematical operations, ensuring computational efficiency while maintaining the ability to capture a wide range of textures.

\begin{figure}[tb]
    \centering    \includegraphics[width=0.49\textwidth]{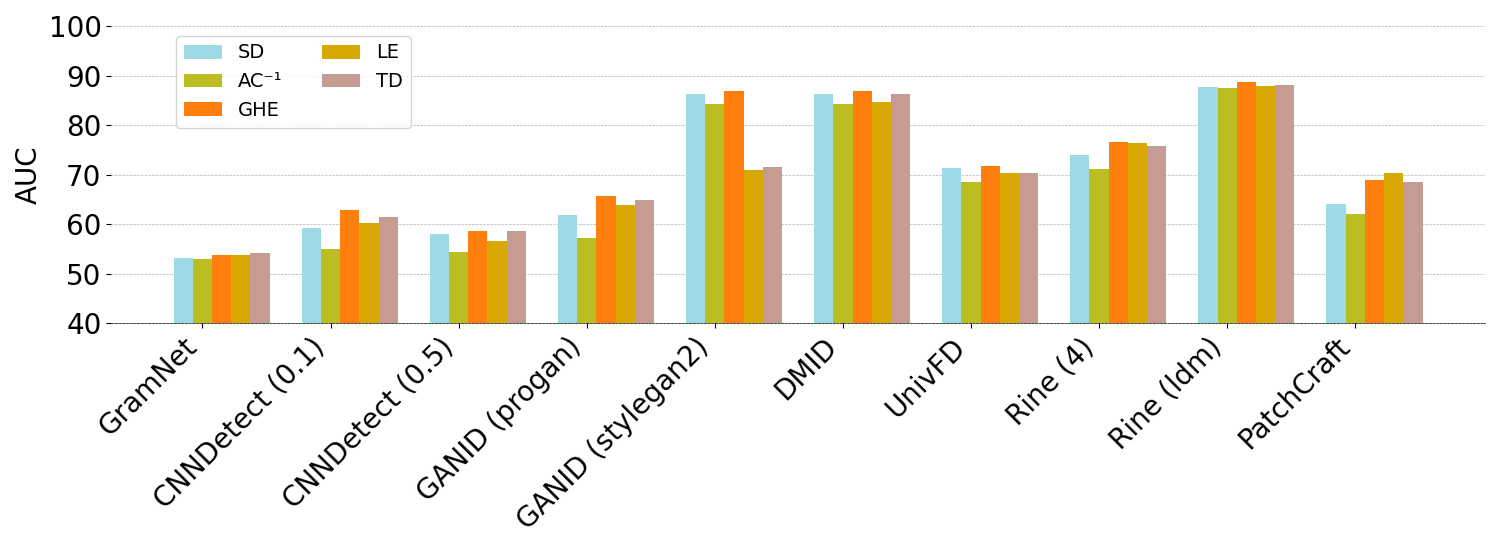}
    
\caption{Performance (AUC) of TextureCrop for different texture selection criteria $\mu$\ across different detectors.}\label{fig:methods}
\end{figure}

TextureCrop (with parameters $\sigma$=$224$, $\omega$=$224\times224$, $\mu$, $\nu$=$10$) selects the top $\nu$=10 crops from the input image based on Global Histogram Entropy (GHE), Local Entropy (LE), Standard Deviation (SD), Inverse Autocorrelation (\(\text{AC}^{-1}\)) and Texture Diversity (TD) of pixel intensities. For all these metrics, higher values indicate more texture-rich regions. Across various models, it appears that the GHE-based method performs better on average and outperforms the LE-, SD-, \(\text{AC}^{-1}\)- and TD-based methods (see~\Cref{fig:methods}).

\subsubsection{Crop Selection Strategy}

To test our hypothesis that texture-rich areas are more important for detection, we also evaluated the bottom 10 crops based on texture information to assess their performance. When contrasting top and bottom crop selections based on the histogram entropy of pixel intensities (GHE), we observe that TextureCrop$\downarrow$($\sigma$=$224$, $\omega$=$224\times224$, $\mu$=$GHE$, $\nu$=$10$) demonstrates significantly worse performance compared to TextureCrop$\uparrow$($\sigma$=224, $\omega$=224$\times$224, $\mu$=GHE, $\nu$=10), where $\uparrow$ and $\downarrow$ denote selection from the top and bottom of the texture distribution, respectively.

\begin{figure}[tb]
    \centering
    \includegraphics[width=0.5\textwidth]{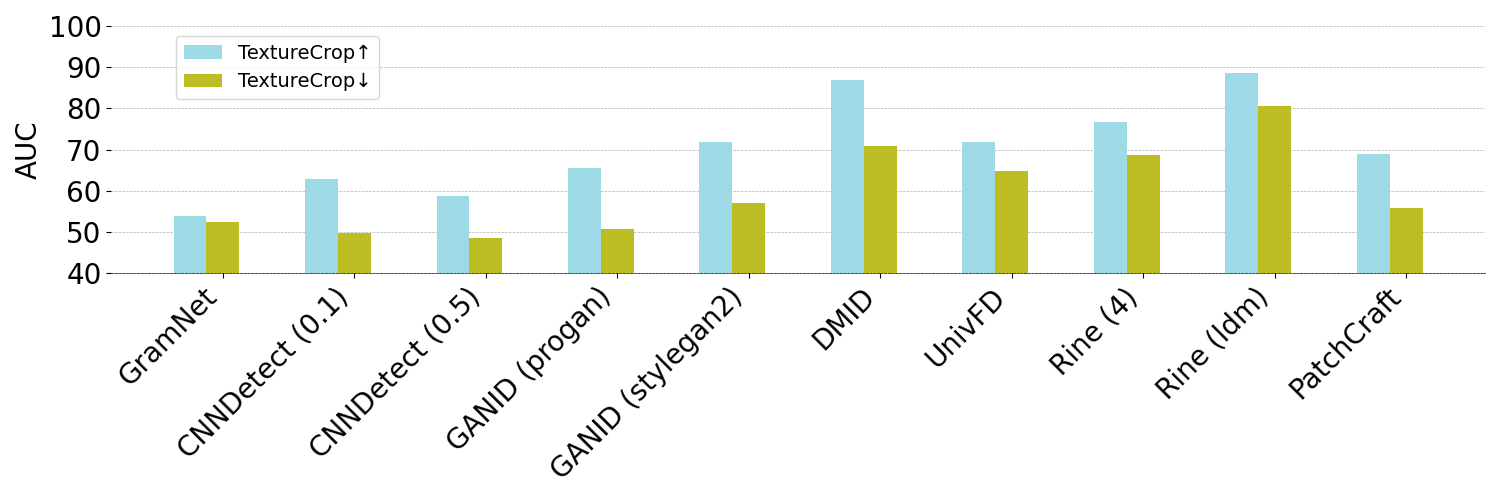}
    \caption{Performance (AUC) of TextureCrop for different crop selection strategies (Top $\uparrow$ vs Bottom $\downarrow$) across different detectors.}        
    \label{fig:position}
\end{figure}

Specifically, it shows an average 6.4\% decrease in BA, a 10.7\% decrease in AP, and a 14.9\% decrease in AUC. This highlights that regions with low histogram entropy do not adequately capture important features for effective detection. These differences in performance are illustrated in~\Cref{fig:position}.

\subsubsection{Aggregation}

\begin{figure}[tb]
    \centering
    \includegraphics[width=0.49\textwidth]{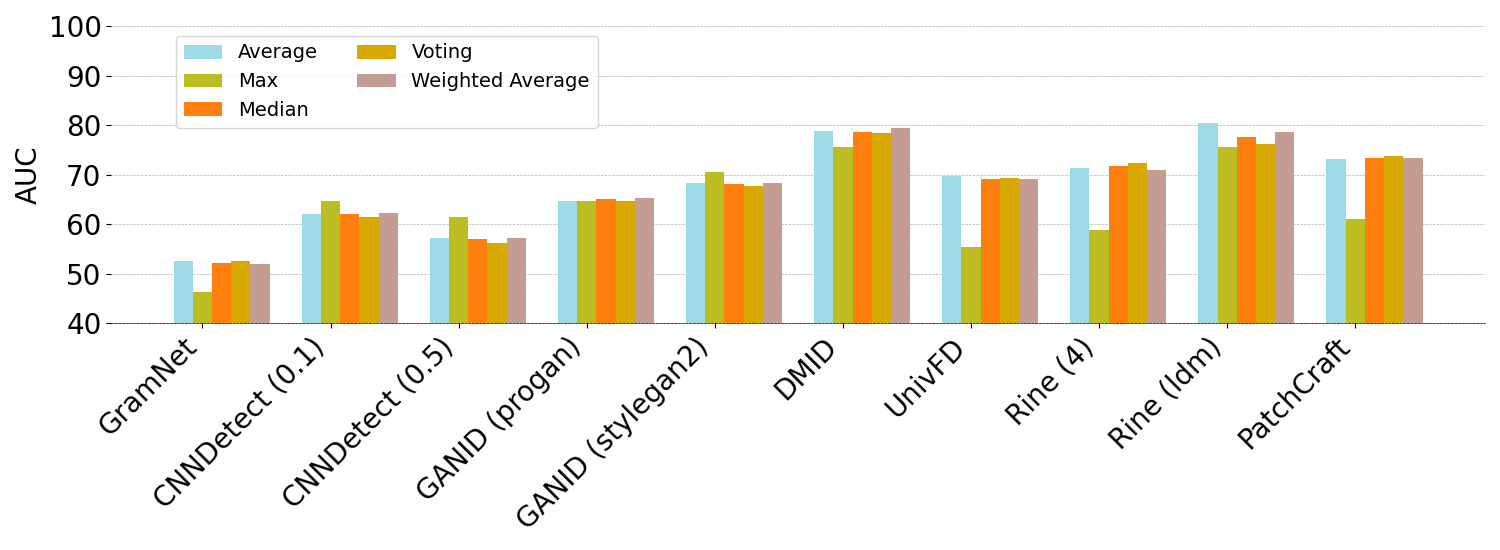}
    \caption{Performance (AUC) of TextureCrop for different aggregation methods $\mathfrak{A}$ across different detectors.}
    \label{fig:fusion}
\end{figure}

As already mentioned earlier, each detector model produces a logit per crop, thus to aggregate the results from multiple crops, we explored the following alternatives for the aggregation function $\mathfrak{A}$: 

\begin{itemize}
    \item \textbf{Majority Voting:} Determines the most frequently predicted class:
    \[
    C_{\text{maj}} = \arg\max_{c} \sum_{i=1}^\nu \mathbb{I}(P_i = c),
    \]
    where $P_i$ is the predicted class for the $i$-th crop, $\mathbb{I}$ is the indicator function, and $c$ iterates over the two possible classes.

    \item \textbf{Max:} Selects the maximum logit value among the crops:
    \[
    s_{\text{max}} = \max_{i} s_i,
    \]
    where $s_i$ is the logit for the $i$-th crop.

    \item \textbf{Median:} Computes the median of the logits from all crops:
    \[
    s_{\text{med}} = \mathop{\mathrm{median}}_{i} (s_i).
    \]
    \item \textbf{Weighted Average:} Computes a weighted average of the logits, with weights determined by the frequency of values in predefined intervals. Let $w_i$ be the weight assigned to the logit $s_i$, and compute:
    \[
    s_{\text{wavg}} = \frac{\sum_{i=1}^\nu w_i s_i}{\sum_{i=1}^\nu w_i}.
    \]
    The weights $w_i$ are proportional to the frequency of $s_i$ in a user-defined interval, with interval length set to 0.1 based on experimental results.
\end{itemize}

The results of these tests are presented in~\Cref{fig:fusion} and show that average, median and weighted average are the most effective aggregation strategies, consistently delivering competitive performance across various detection methods. On the other hand, majority voting and max aggregation appear to be less effective. 

\section{Conclusion}

In this study, we introduced TextureCrop, a pre-processing approach that analyzes texture variability for crop selection. By targeting regions of the image with high texture variability, TextureCrop successfully captures high-frequency artifacts introduced by generative models, resulting in enhanced detection accuracy across various deep learning models. Through our comprehensive experimental study, TextureCrop emerged as a very effective addition for addressing the challenges associated with the task of high-resolution SID. Given its simplicity and ease of implementation, we consider that the majority of SID methods would benefit from its incorporation. 

\section{Acknowledgments}

This work is partially funded by the Horizon 2020 European project vera.ai under grant agreement no. 101070093. 


{\small
\balance
\bibliographystyle{splncs04}
\bibliography{references}

\begin{thebibliography}{10}
\providecommand{\url}[1]{\texttt{#1}}
\providecommand{\urlprefix}{URL }
\providecommand{\doi}[1]{https://doi.org/#1}

\bibitem{adobe2023firefly}
{Adobe Firefly}: \url{https://www.adobe.com/sensei/generative-ai/firefly.html}
  (2024)

\bibitem{amoroso2023parents}
Amoroso, R., Morelli, D., Cornia, M., Baraldi, L., Bimbo, A.D., Cucchiara, R.:
  Parents and children: Distinguishing multimodal deepfakes from natural
  images. arXiv preprint arXiv:2304.00500v1  (2023)

\bibitem{bammey2023synthbuster}
Bammey, Q.: Synthbuster: Towards detection of diffusion model generated images.
  In: IEEE Open Journal of Signal Processing (2023)

\bibitem{betker2023dalle3}
Betker, J., Goh, G., Jing, L., Brooks, T., Wang, J., Li, L., Ouyang, L.,
  Zhuang, J., Lee, J., Guo, Y., Manassra, W., Dhariwal, P., Chu, C., Jiao, Y.:
  https://openai.com/dall-e-3 (2023)

\bibitem{brock2019biggan}
Brock, A., Donahue, J., Simonyan, K.: Large scale gan training for high
  fidelity natural image synthesis. In: ICLR (2019)

\bibitem{chai2020patchforensics}
Chai, L., Bau, D., Lim, S.N., Isola, P.: What makes fake images detectable?
  understanding properties that generalize. In: ECCV (2020)

\bibitem{chen2018sitd}
Chen, C., Chen, Q., Xu, J., Koltun, V.: Learning to see in the dark. In:
  IEEE/CVF Conference on Computer Vision and Pattern Recognition (CVPR) (2020)

\bibitem{chen2022ost}
Chen, L., Zhang, Y., Song, Y., Wang, J., Liu, L.: Ost: Improving generalization
  of deepfake detection via one-shot test-time training. In: NeurIPS (2022)

\bibitem{chen2023twigmadatasetaigeneratedimages}
Chen, Y., Zou, J.: Twigma: A dataset of ai-generated images with metadata from
  twitter (2023), \url{https://arxiv.org/abs/2306.08310}

\bibitem{choi2018stargan}
Choi, Y., Choi, M., Kim, M., Ha, J.W., Kim, S., Choo, J.: Stargan: Unified
  generative adversarial networks for multi-domain image-to-image translation.
  In: CVPR (2018)

\bibitem{corvi2022detection}
Corvi, R., Cozzolino, D., Zingarini, G., Poggi, G., Nagano, K., Verdoliva, L.:
  On the detection of synthetic images generated by diffusion models. In:
  ICASSP 2023-2023 IEEE International Conference on Acoustics, Speech and
  Signal Processing (ICASSP). pp.~1--5 (2023)

\bibitem{cozzolino2023raising}
Cozzolino, D., Poggi, G., Corvi, R., Nießner, M., Verdoliva, L.: Raising the
  bar of ai-generated image detection with clip. arXiv preprint
  arXiv:2312.00195  (2023)

\bibitem{croitoru2022diffusion}
Croitoru, F.A., Hondru, V., Ionescu, R.T., Shah, M.: Diffusion models in
  vision: A survey. IEEE Transactions on Pattern Analysis and Machine
  Intelligence,  \textbf{14} (2022)

\bibitem{dai2019san}
Dai, T., Cai, J., Zhang, Y., Xia, S.T., Zhang, L.: Second-order attention
  network for single image super-resolution. In: IEEE/CVF Conference on
  Computer Vision and Pattern Recognition (CVPR). pp. 11065--11074 (2019)

\bibitem{ductien2015raise}
Dang-Nguyen, D.T., Pasquini, C., Conotter, V., Boato, G.: Raise: a raw images
  dataset for digital image forensics. In: Proceedings of the 6th ACM
  Multimedia Systems Conference (2015)

\bibitem{dhariwal2021diffusion}
Dhariwal, P., Nichol, A.: Diffusion models beat gans on image synthesis. In:
  Advances in Neural Information Processing Systems (2021)

\bibitem{dosovitskiy2020transformers}
Dosovitskiy, A., Beyer, L., Kolesnikov, A., Weissenborn, D., Zhai, X.,
  Unterthiner, T., Dehghani, M., Minderer, M., Heigold, G., Gelly, S.,
  Uszkoreit, J., Houlsby, N.: An image is worth 16x16 words: Transformers for
  image recognition at scale. arXiv preprint arXiv:2010.11929  (2020)

\bibitem{frank2020leveraging}
Frank, J., Eisenhofer, T., Schönherr, L., Fischer, A., Kolossa, D., Holz, T.:
  Leveraging frequency analysis for deep fake image recognition. In:
  Proceedings of the 37th International Conference on Machine Learning, PMLR
  119 (2020)

\bibitem{goodfellow2014gan}
Goodfellow, I.J., Pouget-Abadie, J., Mirza, M., Xu, B., Warde-Farley, D.,
  Ozair, S., Courville, A., Bengio, Y.: Generative adversarial nets. In:
  Advances in Neural Information Processing Systems (NIPS) (2014)

\bibitem{gragnaniello2021gan}
Gragnaniello, D., Cozzolino, D., Marra, F., Poggi, G., Verdoliva, L.: Are gan
  generated images easy to detect? a critical analysis of the state-of-the-art.
  In: IEEE ICME (2021)

\bibitem{kaiming2016resnet}
He, K., Zhang, X., Ren, S., Sun, J.: Deep residual learning for image
  recognition. In: Proceedings of the IEEE/CVF Conference on Computer Vision
  and Pattern Recognition (2016)

\bibitem{ho2020diffusion}
Ho, J., Jain, A., Abbeel, P.: Denoising diffusion probabilistic models. In:
  Advances in Neural Information Processing Systems. vol.~33, pp. 6840--6851
  (2020)

\bibitem{jabbar2020survey}
Jabbar, A., Li, X., Omar, B.: A survey on generative adversarial networks:
  Variants, applications, and training. ACM Computing Surveys  \textbf{54},
  1–49 (2020)

\bibitem{jeong2022frepgan}
Jeong, Y., Kim, D., Ro, Y., Choi, J.: Frepgan: Robust deepfake detection using
  frequency- level perturbations. In: AAAI (2022)

\bibitem{chen2024single}
Jiaxuan~Chen, Jieteng~Yao, L.N.: A single simple patch is all you need for
  ai-generated image detection. arXiv preprint arXiv:2402.01123  (2024)

\bibitem{ju2022fusing}
Ju, Y., Jia, S., Ke, L., Xue, H., Nagano, K., Lyu, S.: Fusing global and local
  features for generalized ai-synthesized image detection. In: IEEE ICIP (2022)

\bibitem{karras2018progan}
Karras, T., Aila, T., Laine, S., Lehtinen, J.: Progressive growing of gans for
  improved quality, stability, and variation. In: ICLR (2018)

\bibitem{karras2021alias}
Karras, T., Aittala, M., Laine, S., Härkönen, E., Hellsten, J., Lehtinen, J.,
  Aila, T.: Alias-free generative adversarial networks. In: 35th Conference on
  Neural Information Processing Systems (NeurIPS). (2021)

\bibitem{karras2019stylegan}
Karras, T., Laine, S., Aila, T.: A style-based generator architecture for
  generative adversarial networks. In: CVPR (2019)

\bibitem{karras2020analyzing}
Karras, T., Laine, S., Aittala, M., Hellsten, J., Lehtinen, J., Aila, T.:
  Analyzing and improving the image quality of stylegan. In: CVPR (2020)

\bibitem{koutlis2024leveraging}
Koutlis, C., Papadopoulos, S.: Leveraging representations from intermediate
  encoder-blocks for synthetic image detection. arXiv preprint arXiv:2402.1909
  (2024)

\bibitem{openimages}
Kuznetsova, A., Rom, H., Alldrin, N., Uijlings, J., Ivan~Krasin, J.P.T.,
  Kamali, S., Popov, S., Malloci, M., Kolesnikov, A., Duerig, T., Ferrari, V.:
  The open images dataset v4: Unified image classification, object detection,
  and visual relationship detection at scale. CoRR  \textbf{abs/1811.00982}
  (2018)

\bibitem{liu2020global}
Liu, Z., Qi, X., Torr, P.: Global texture enhancement for fake face detection
  in the wild. In: CVPR (2020)

\bibitem{luFake2M}
Lu, Z., Huang, D., Bai, L., Qu, J., Wu, C., Liu, X., Ouyang, W.: Seeing is not
  always believing: Benchmarking human and model perception of ai-generated
  images. In: Thirty-seventh Conference on Neural Information Processing
  Systems Datasets and Benchmarks Track (2023)

\bibitem{midjourney2023}
{Midjourney}: \url{https://www.midjourney.com} (2024)

\bibitem{mirza2014cgan}
Mirza, M., Osindero, S.: Conditional generative adversarial nets. In: Advances
  in Neural Information Processing Systems (NIPS) (2014)

\bibitem{nichol2021glide}
Nichol, A., Dhariwal, P., Ramesh, A., Shyam, P., Mishkin, P., McGrew, B.,
  Sutskever, I., Chen, M.: Glide: Towards photorealistic image generation and
  editing with text-guided diffusion models. In: ICML (2021)

\bibitem{ojha2023universal}
Ojha, U., Li, Y., Lee, Y.J.: Towards universal fake image detectors that
  generalize across generative models. In: CVPR. pp. 24480--24489 (2023)

\bibitem{papa2023ontheuse}
Papa, L., Faiella, L., Corvitto, L., Maiano, L., Amerini, I.: On the use of
  stable diffusion for creating realistic faces: From generation to detection.
  In: 11th International Workshop on Biometrics and Forensics (IWBF) (2023)

\bibitem{park2019gaugan}
Park, T., Liu, M.Y., Wang, T.C., Zhu, J.Y.: Semantic image synthesis with
  spatially-adaptive normalization. In: CVPR (2019)

\bibitem{podell2023sdxl}
Podell, D., English, Z., Lacey, K., Blattmann, A., Dockhorn, T., Müller, J.,
  Penna, J., Rombach, R.: Sdxl: Improving latent diffusion models for
  high-resolution image synthesis. arXiv preprint arXiv:2307.01952  (2023)

\bibitem{quian2020thinking}
Qian, Y., Yin, G., Sheng, L., Chen, Z., Shao, J.: Thinking in frequency: Face
  forgery detection by mining frequency-aware clues. In: ECCV (2020)

\bibitem{radford2021clip}
Radford, A., Kim, J., Hallacy, C., Ramesh, A., Goh, G., Agarwal, S., Sastry,
  G., Askell, A., Mishkin, P., et~al., J.C.: Learning transferable visual
  models from natural language supervision. In: ICML. pp. 8748--8763 (2021)

\bibitem{radford2016dcgan}
Radford, A., Metz, L., Chintala, S.: Unsupervised representation learning with
  deep convolutional generative adversarial networks. In: ICLR (2016)

\bibitem{ramesh2022dalle2}
Ramesh, A., Dhariwal, P., Nichol, A., Chu, C., Chen, M.: Hierarchical
  text-conditional image generation with clip latents. arXiv preprint
  arXiv:2204.06125  (2022)

\bibitem{ren2016fasterrcnnrealtimeobject}
Ren, S., He, K., Girshick, R., Sun, J.: Faster r-cnn: Towards real-time object
  detection with region proposal networks (2016),
  \url{https://arxiv.org/abs/1506.01497}

\bibitem{rombach2022latent}
Rombach, R., Blattmann, A., Lorenz, D., Esser, P., Ommer, B.: High-resolution
  image synthesis with latent diffusion models. In: Proceedings of the IEEE/CVF
  Conference on Computer Vision and Pattern Recognition. pp. 10684--10695
  (2022)

\bibitem{rombach2022stable}
Rombach, R., Blattmann, A., Lorenz, D., Esser, P., Ommer, B.: Stable diffusion
  (2022)

\bibitem{schinas2024sidbenchpythonframeworkreliably}
Schinas, M., Papadopoulos, S.: Sidbench: A python framework for reliably
  assessing synthetic image detection methods (2024),
  \url{https://arxiv.org/abs/2404.18552}

\bibitem{sezgin2004new}
Sezgin, M., Sankur, B.: A new method for gray-level picture thresholding using
  the entropy of the histogram. Journal of Electronic Imaging  \textbf{13},
  146--153 (2004)

\bibitem{sha2023defake}
Sha, Z., Li, Z., Yu, N., Zhang, Y.: Defake: Detection and attribution of fake
  images generated by text-to-image diffusion models. In: ACM SIGSAC Conference
  on Computer and Communications Security. pp. 3418--3432 (2023)

\bibitem{song2021score}
Song, Y., Sohl-Dickstein, J., Kingma, D.P., Kumar, A., Ermon, S., Poole, B.:
  Score-based generative modeling through stochastic differential equations.
  In: Proceedings of the International Conference on Learning Representations
  (2021)

\bibitem{wang2020cnngenerated}
Wang, S.Y., Wang, O., Zhang, R., Owens, A., Efros, A.A.: Cnn-generated images
  are surprisingly easy to spot... for now. In: CVPR. pp. 8695--8704 (2020)

\bibitem{wang2021generative}
Wang, Z., She, Q., Ward, T.: Generative adversarial networks in computer
  vision: A survey and taxonomy. ACM Computing Surveys  (2021)

\bibitem{whichfaceisreal}
{Which face is real?}: \url{http://www.whichfaceisreal.com}

\bibitem{yang2023diffusion}
Yang, L., Zhang, Z., Song, Y., Hong, S., Xu, R., Zhao, Y., Zhang, W., Cui, B.,
  Yang, M.H.: Diffusion models: A comprehensive survey of methods and
  applications. ACM Computing Surveys  \textbf{56},  1–39 (2023)

\bibitem{yu2015lsun}
Yu, F., Zhang, Y., Song, S., Seff, A., Xiao, J.: Lsun: Construction of a
  large-scale image dataset using deep learning with humans in the loop. arXiv
  preprint arXiv:1506.03365  (2015)

\bibitem{zhong2024patchcraft}
Zhong, N., Xu, Y., Li, S., Qian, Z., Zhang, X.: Patchcraft: Exploring texture
  patch for efficient ai-generated image detection. arXiv preprint
  arXiv:2311.12397  (2024)

\bibitem{zhu2017cyclegan}
Zhu, J.Y., Park, T., Isola, P., Efros, A.A.: Unpaired image-to-image
  translation using cycle-consistent adversarial networks. In: ICCV (2017)

\end{thebibliography}
}

\end{document}